\documentclass[11pt,a4paper]{article}
\usepackage[hyperref]{naaclhlt2019}
\usepackage{times}
\usepackage{latexsym}
\usepackage{graphicx}
\usepackage{multirow}
\usepackage{booktabs}

\usepackage{url}
\aclfinalcopy 
\def\aclpaperid{1974} 

\title{Cross-Lingual Transfer Learning for Multilingual Task Oriented Dialog}

\author{Sebastian Schuster\thanks{\ \ \ Work carried out during an internship at Facebook.}\\
  Stanford Linguistics\\
  \texttt{sebschu@stanford.edu} \\
   \And
   Sonal Gupta \\
   Facebook Conversational AI \\
   \texttt{sonalgupta@fb.com} \\
   \AND 
    Rushin Shah  \\
   Facebook Conversational AI \\
   \texttt{rushinshah@fb.com} \\ \\
   \And
      Mike Lewis \\
   Facebook AI Research \\
   \texttt{mikelewis@fb.com} 
}

\begin{document}

\maketitle

\begin{abstract}
One of the first steps in the utterance interpretation pipeline of many task-oriented conversational AI systems is to identify user intents and the corresponding slots. Since data collection for machine learning models for this task is time-consuming, it is desirable to make use of existing data in a high-resource language to train models in low-resource languages. However, development of such models has largely been hindered by the lack of multilingual training data. In this paper, we present a new data set of 57k annotated utterances in English (43k), Spanish (8.6k) and Thai (5k) across the domains weather, alarm, and reminder. We use this data set to evaluate three different cross-lingual transfer methods: (1) translating the training data, (2) using cross-lingual pre-trained embeddings, and (3) a novel method of using a multilingual machine translation encoder as contextual word representations. We find that given several hundred training examples in the the target language, the latter two methods outperform translating the training data. Further, in very low-resource settings, multilingual contextual word representations give better results than using cross-lingual static embeddings. We also compare the cross-lingual methods to using monolingual resources in the form of contextual ELMo representations and find that given just small amounts of target language data, this method outperforms all cross-lingual methods, which highlights the need for more sophisticated cross-lingual methods.

\end{abstract}

\section{Introduction}

One of the first steps in many conversational AI systems that are used to parse utterances in personal assistants is the identification of what the user intends to do (the {\it intent}) as well as the arguments of the intent (the {\it slots})~\cite{mesnil2013,Liu2016AttentionBasedRN}. For example, for a request such as {\it Set an alarm for tomorrow at 7am}, a first step in fulfilling such a request is to identify that the user's intent is to set an alarm and that the required time argument of the request is expressed by the phrase {\it tomorrow at 7am}.

Given these properties of the task, the problem can be stated as a joint sentence classification (for intent classification) and sequence labeling (for slot detection) task and therefore naturally lends itself to using a biLSTM-CRF sequence labeling model~\cite{lamplener, Vu2016} where the biLSTM layer is also used as the input for a projection layer for intent detection.

These models are very powerful and given enough training data, they achieve very high accuracy on the intent classification as well as the slot detection task. However, given the requirement of large amounts of labeled training data, developing a conversational AI system for many new languages is a very resource-intensive task and clearly not feasible to be done for the more than 6,500  languages that are currently spoken around the world.  

\begin{table*}[]
\small
    \centering
    \begin{tabular}{@{}l|clc|c|c@{}}
                  & \multicolumn{3}{c|}{\textbf{Number of utterances}}                                              & \multirow{2}{*}{\textbf{Intent types}} & \multirow{2}{*}{\textbf{Slot types}} \\
\textbf{Domain}   & \textbf{English}   & \multicolumn{1}{c}{\textbf{Spanish}} & \multicolumn{1}{c|}{\textbf{Thai}} &                                        &                                      \\ \midrule
\textbf{Alarm}    & 9,282/1,309/2,621  & 1,184/691/1,011                      & 777/439/597                        & 6                                      & 2                                    \\
\textbf{Reminder} & 6,900/943/1,960    & 1,207/647/1,005                      & 578/336/442                        & 3                                      & 6                                    \\
\textbf{Weather}  & 14,339/1,929/4,040 & 1,226/645/1,027                      & 801/460/653                        & 3                                      & 5 \\ \midrule
\textbf{\textit{Total}}  & 30,521/4,181/8,621 & 3,617/1,983/3,043                      & 2,156/1,235/1,692                        &                      12                 & 11                                    \\ 
\end{tabular}

    \caption{Summary statistics of the data set. The three values for the number of utterances correspond to the number of utterances in the training, development, and test splits. Note that the slot type \textit{datetime} is shared across all three domains and therefore the total number of slot types is only 11.}
    \label{tbl:data-stats}
\end{table*}

For this reason, one would like to make use of methods that enable transfer learning from a high-resource language to a low-resource language. However, the development of sophisticated cross-lingual transfer methods for intent detection and slot filling has so far been hindered by the lack of multilingual data sets that have been annotated according to the same guidelines.\footnote{\citet{Upadhyay2018} collected such a dataset but to the best of our knowledge, their data is not publicly available.} In this work, we therefore present a novel data set containing a large number of English utterances (the high-resource data) as well as a smaller set of utterances in Spanish and in Thai (the low-resource data), which were annotated according to the same annotation scheme. This data allows the systematic investigation of cross-lingual transfer learning methods from high-resource languages to low-resource languages.

We use this data set to explore different strategies to make use of training data from a high-resource language to improve intent and slot detection models for other languages. We investigate two previously proposed strategies for cross-lingual transfer, namely using cross-lingual pre-trained embeddings (XLU embeddings; see \citealp{Ruder2017} for a review) as well as automatically translating the English training data to the target language. Further, we present a novel technique that uses a bidirectional neural machine translation encoder as cross-lingual contextual word representations. We compare the cross-lingual transfer methods to models that are only trained on the target language data.

Across the two languages and the various transfer methods, we find 
that joint training on the high-resource and the low-resource target language improves results on the target language.
We further find that the optimal choice of transfer method depends on the size of the available training data in the target language: 
In the zero-shot case where no target language data is available, translating the training data gives the best results. However, 
if a small amount of training data is available, we find that jointly training on the high-resource and low-resource data works
better than training on translated data.

We release the data at \url{https://fb.me/multilingual_task_oriented_data}.
\section{Data}

We originally collected a data set of around 43,000 English utterances across the domains \textsc{alarm}, \textsc{reminder}, and \textsc{weather}. Data collection proceeded in three steps. First, native English speakers were asked to produce utterances for each intent, e.g., provide examples of how they would ask for the weather. In a second step, two annotators would label the intent and the spans corresponding to slots for each utterance. As a third step, if annotators disagreed on the annotation of an utterance, a third annotator who corresponded with the authors of the guidelines adjudicated between the two annotations.

For the Spanish and Thai data, native speakers of the target language translated a sample of the English utterances. These translated utterances were then also annotated by two annotators. For Spanish, if annotators disagreed, a third annotator who was bilingual in Spanish and English adjudicated these disagreements in communication with the guideline authors. Unfortunately, for Thai, we did not have a bilingual speaker available and hence we decided to discard all utterances for which the annotators disagreed. We hope to rectify this for future data collection efforts.

We believe this data presents a great opportunity to investigate cross-lingual semantic models and to the best of our knowledge, this is the first parallel data set for a word tagging task that has been annotated according to the same guidelines across multiple languages.

Table~\ref{tbl:data-stats} contains several summary statistics of the data set. Note that the percentage of training examples as compared to development and test examples is much higher for the English data than for the Thai and Spanish data. We decided for a more even split for the latter two languages so that we had a sufficiently large data set for model selection and evaluation.

\section{NLU models}

\begin{figure}
    \centering
    \includegraphics[width=\columnwidth]{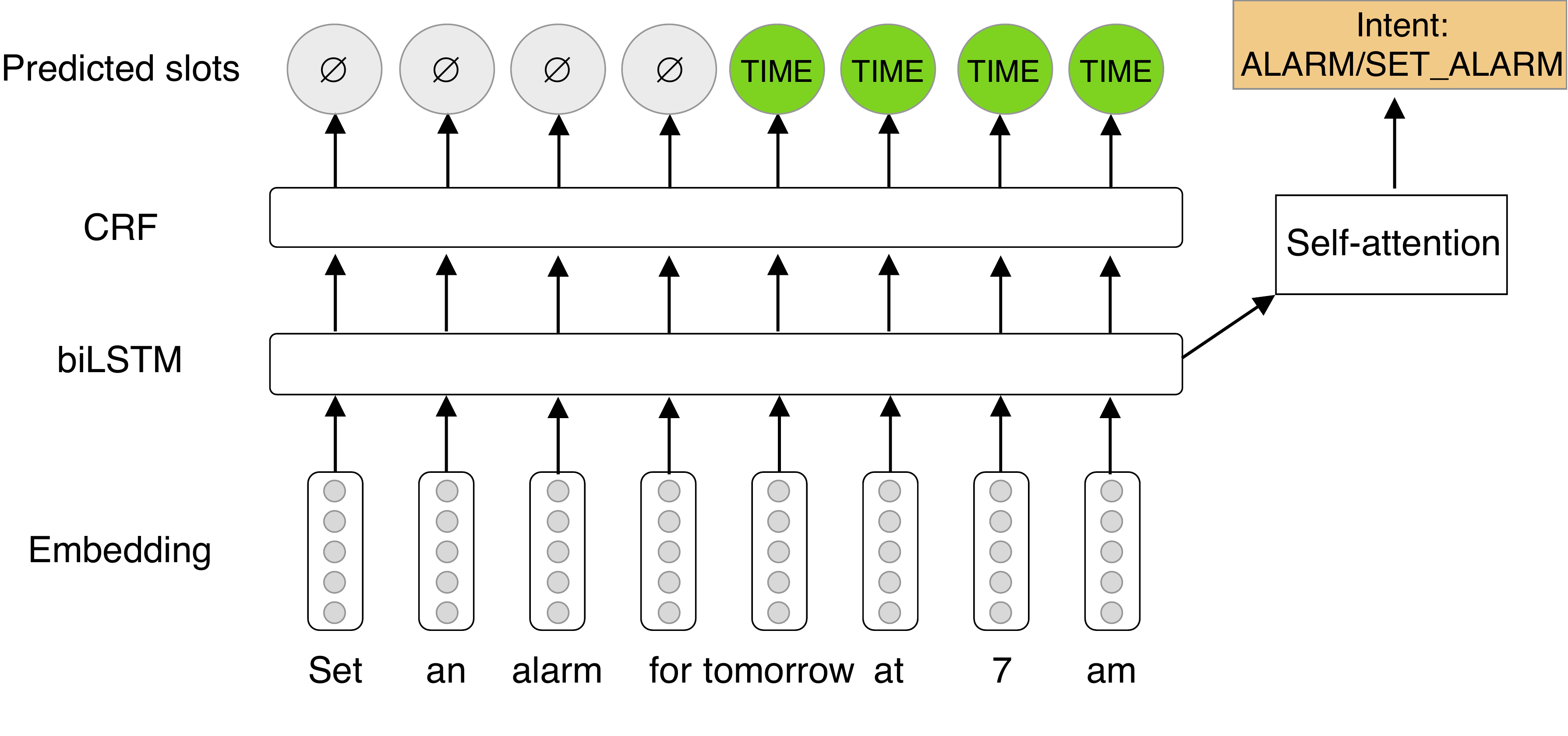} 
    \caption{Slot and intent model architecture. Word embeddings are passed through a biLSTM layer which is shared across the slot detection and intent prediction tasks.}
    \label{fig:model-architecture}
\end{figure}

The intent detection and slot-filling model consists of two parts: It first uses a sentence classification model to identify the domain of the user utterance (in our case, \textsc{alarm}, \textsc{reminder}, or \textsc{weather}), and then uses a domain-specific model to jointly predict the intent and slots. Figure~\ref{fig:model-architecture} shows the basic architecture of the joint intent-slot prediction model. It first embeds the utterance using an embedding matrix and then passes the word vectors to a biLSTM layer. For intent classification, we use a self-attention layer \citep{Lin2017} over the hidden states of the biLSTM as input to a softmax projection layer; for slot detection, we pass for each word the concatenation of the forward and backward hidden states through a softmax layer, and then pass the resulting label probability vectors through a CRF layer for final predictions.

In our experiments, we vary how the tokens are embedded:
\begin{itemize}
    \item \textbf{Zero embeddings}: We train the parameters of a 0-initialized embedding matrix that contains each word that appears in the training data.
    \item \textbf{XLU embeddings}: We embed the tokens through lookup in a pre-trained cross-lingual embedding matrix and concatenate these embeddings with tuned zero embeddings. Here, we follow \citet{Dozat2017} in having a fixed pre-trained embedding matrix combined with tuneable zero-embeddings.
    \item \textbf{Encoder embeddings}: We embed tokens by passing the entire utterance through a pre-trained biLSTM sentence encoder and using the hidden states of the top layer as input. We keep the parameters of the pre-trained encoder fixed and concatenate them with tuneable zero-embeddings. (See Section~\ref{sec:mt-encoder} for a detailed description of the different encoders.)
\end{itemize}

\begin{table*}
   \small
    \centering
    \begin{tabular}{l |c | c | c | c | c}
        \textbf{Spanish}& Epoch &es$\rightarrow$en & en$\rightarrow$es & es$\rightarrow$es & en$\rightarrow$en\\ \midrule
        CoVe (unidirectional)  &81 & 8.50 & - & - & -  \\
        Mult. CoVe  & 98 & 8.27 & 6.90 & - & -  \\
        Mult. CoVe  + autoencoder & 282 & 9.15 & 7.29 & 1.15 & 1.14 \\ \midrule \midrule
        \textbf{Thai}& Epoch & th$\rightarrow$en & en$\rightarrow$th & th$\rightarrow$th & en$\rightarrow$en\\ \midrule
        CoVe (unidirectional) & 12 & 13.06  & - & - & -  \\
        Mult. CoVe  & 35 & 12.73 & 17.00  & - & -  \\
        Mult. CoVe  + autoencoder & 92 & 11.76 & 16.31  & 1.12 & 1.13   
    \end{tabular}
    \caption{Perplexities on validation set for different encoder models for the Spanish-English and Thai-English language pairs. A hyphen means that an encoder was not trained for the corresponding language pair.}
    \label{tbl:results-nmt}
\end{table*}

\section{Encoder models}
\label{sec:mt-encoder}
As mentioned in the previous section, some of our models use a pre-trained biLSTM encoder to generate contextual word embeddings. In all our experiments, we use a bidirectional LSTM encoder with two layers. Overall, we compare three strategies for training these encoders. The motivation for choosing these strategies is to investigate whether there is a benefit of using multilingual embeddings.
\begin{itemize}
    \item \textbf{CoVe}: Following \citet{McCann2017}, we train a neural machine translation model to translate from the low-resource language (Spanish or Thai) to English.
    \item \textbf{Multilingual CoVe}: We train a neural machine translation model to translate from the  low-resource language to English and from English to the  low-resource language. We encode the translation direction using target language-specific start tokens in the decoder \cite{YuK2018}. In this model, the encoder does not have access to the target language and therefore we expect it to learn  to encode phrases with similar meanings into similar vector spaces across languages. 
    \item \textbf{Multilingual CoVe w/ autoencoder:} We train a bidirectional neural machine translation model and combine it with an auto-encoder objective. For the language pair Spanish-English, that means given a Spanish input sentence we train the model to generate either an English translation or to reproduce the Spanish sentence depending on the start token in the decoder. Likewise, given an English sentence, we train the model to output either a Spanish translation or to reproduce the English sentence depending on the start token in the decoder. The motivation for this approach is that using the joint translation and autoencoder objective might lead to more general representations since the decoder has to be capable to output sentences in either language independent of what the source language was, and unlike in the previous model the source language does not determine the target language. We train an analogous model for the Thai-English language pair.
\end{itemize}

\noindent For Spanish, for which pre-trained ELMo \citep{Peters2018} embeddings are available, we also evaluate the use of the ELMo embeddings by \citet{Che2018,Fares2017}.
Note that the ELMo encoder and the CoVe encoder are trained to encode only the low-resource language and therefore neither of them are multilingual encoders.

\paragraph{Implementation details} We train all models using a wrapper around the fairseq \cite{Gehring2016,Gehring2017} sequence-to-sequence models. We use 300d randomly initialized word vectors as input to the first embedding layer. Each direction in each hidden layer has 512 dimensions which results in a total encoder output dimension  of 1024. For the machine translation models, we further use dot-product attention \citep{Luong2015} and to improve efficiency, we limit the output space of the softmax to 30 translation candidates as determined by word alignments as well as the 2,000 most frequent words \cite{Lhostis2016}.

\paragraph{Data} For the Spanish models, we use two copies\footnote{We upsample the Europarl (for Spanish) and IWSLT (for Thai) data since these data sets are presumably of higher quality than the largerly automatically mined Paracrawl and OpenSubtitles data.} of Europarl v7 \cite{Koehn2005}, every eighth sentences of the Paracrawl data,\footnote{https://paracrawl.eu, the version that was used in the WMT 2018 task} and the newstest2008-2011 data. For model selection, we use the newstest2012-2013 data. For the Thai models, we use 10 copies of the IWSTL training data \cite{Cettolo2012} as well as the OpenSubtitles data \cite{Lison2018} for training, and the IWSTL development and test data for model selection. We use the 20,000 most common words in the training data as the vocabulary. For the multilingual models, we take the union of the vocabulary from both languages. We tokenize the data using an in-house rule-based (for English and Spanish) and dictionary-based (for Thai) tokenizer. We further lowercase all data and remove all duplicates within a data set. We discard all sentences whose length exceeds 100 tokens.

\paragraph{Training details} We train the models using stochastic gradient descent with an initial learning rate of 0.5. We decrease the learning rate by 1\% after an epoch whenever perplexity on the validation data is higher than for the epoch with the lowest perplexity. We train all models for up to 100 epochs, except for the Spanish bidirectional MT model with an autoencoder which we trained for 300 epochs since it took considerably longer to converge. For multilingual models, we choose the model that has the lowest average perplexity on both translation tasks.

Table~\ref{tbl:results-nmt} shows the perplexities for the different models. In general, the translation perplexities are very similar independent of whether we train a unidirectional MT system or a bidirectional MT system, except for the Spanish bidirectional MT model with an autoencoder which even after 300 epochs still yields higher perplexities on the validation data than the other translation models.\footnote{We hypothesize that the slow convergence as well as the lower performance might be caused by the fact that the sentences in the Spanish-English parallel data are much longer than in the Thai-English data which might make it harder to learn good universal sentence representations.}

\begin{table*}[t]
   \small
    \centering
    \begin{tabular}{l | l | c|c|c|c}
        \toprule \\
        \textbf{English} & Embedding type & Exact match & Domain acc. & Intent acc. & Slot F1   \\ \midrule
          Target only & - & 90.91 & - & 99.11 & 94.81 
\\   \midrule \midrule
        \textbf{Spanish} & Embedding type & Exact match & Domain acc. & Intent acc. & Slot F1   \\ \midrule
          Target only & - & 72.94 & 99.43 & 97.26 & 80.95 \\
          Target only & XLU embeddings&	72.90 & 	99.47 & 96.90&	80.99  \\
          Target only & CoVe & 73.93	 & 99.52 & 97.43 &		81.51 \\
          Target only & Mult. CoVe  &  74.13	 & 99.55 & 97.61 &		81.64 \\
          Target only & Mult. CoVe  + auto & 73.05 &	99.51 & 97.13 &		81.22 \\
          Target only & ELMo &  	74.81 &	99.53 & 96.64 & 82.96 
 \\ \midrule
         Translate train & - & 72.49 &	\textbf{99.65} & \textbf{98.47} &		80.60 \\ \midrule
         Cross-lingual & XLU embeddings & {75.39} 		& 99.52 & 97.68 &	{83.00}
 \\ 
         Cross-lingual & CoVe & 75.17  & 99.55 & 97.81 &	 82.55 \\ 
         Cross-lingual & Mult. CoVe & 75.20	& 99.56 & 97.82 &		82.49 \\ 
         Cross-lingual & Mult. CoVe + auto &74.68 & 99.59 & 97.90 &			82.13 \\ 
          Cross-lingual & ELMo & \textbf{75.96}  & 99.47	 & 97.51 &		\textbf{83.38}\\      \midrule \midrule
        \textbf{Thai} & Embedding type & Exact match & Domain acc. & Intent acc. & Slot F1 \\ \midrule
          Target only & - & 79.80	& 99.31 &  95.13	& 87.26 \\
          Target only & CoVe & 84.84 &	99.36 & 96.60 &	90.63 \\
          Target only & Mult. CoVe & 84.66 & 99.37 & 96.75 &		90.20 \\
          Target only & Mult. CoVe + auto & 84.79 &	\textbf{99.41} & 96.59 &	90.51 \\ \midrule
         Translate train & - & 73.37 & 99.37 & \textbf{97.41} &			80.38  \\ \midrule
         Cross-lingual & CoVe & 	84.49 & 99.29 & 96.87 &		90.60 \\ 
         Cross-lingual & Mult. CoVe & 85.76 & 99.39 & 96.98 & 91.22 \\ 
         Cross-lingual & Mult. CoVe  + auto & \textbf{86.12} & 99.33 & 96.87 &		\textbf{91.51}
    \end{tabular}
    \vspace{1em}
    \caption{Results using the full training data averaged over 5 training runs. The \textit{translate train} models are trained on the union of translated English and target language data; the \textit{cross-lingual} models are trained on English and target language data.  }
    \label{tbl:full-results}
\end{table*}

\section{Cross-lingual learning}

In our first set of experiments, we explore the following baselines and strategies for training models in Spanish and Thai given  a large amount of English training data and a small amount of Spanish and Thai training data.
\begin{itemize}
    \item \textbf{Target only}: Using only the low-resource target language data.
     \item \textbf{Target only with encoder embeddings}: Using only the low-resource language training data and using pre-trained encoder embeddings.
    \item \textbf{Translate train}: Combining the target training data with the English data that has been automatically translated to the target language. The slot annotations are projected via the attention weights \cite{Yarowsky2001}. We translate the data using the Facebook neural machine translation system.
    \item \textbf{Cross-lingual with XLU embeddings}: Joint training on the English and target language data with pre-trained MUSE \cite{Conneau2017} cross-lingual embeddings. Since MUSE embeddings are not available for Thai, we only evaluate this method for Spanish.
    \item \textbf{Cross-lingual with encoder embeddings}: Joint training on the English and target language data using pre-trained encoder embeddings.
\end{itemize}

\begin {table*}
   \small
\centering
\begin{tabular}{l | l | c|c|c|c}
        \textbf{Spanish} & Embedding type &  Exact match & Domain acc. & Intent acc. & Slot F1 \\ \midrule
         Translate train & - &	\textbf{54.95} &	\textbf{88.70} & \textbf{85.39} &	\textbf{72.87}  \\ \midrule 
         Cross-lingual & - & 0.63	& 37.74 & 36.17 &		5.50 \\ 
         Cross-lingual & XLU embeddings & 4.01 & 38.24 &  36.94 &			17.50 \\
         Cross-lingual & CoVe & 	1.37 & 39.42 & 37.13 &		5.35  \\
         Cross-lingual & mult. CoVe & {10.56} & {59.29} & {53.34} &			{22.50} \\
         Cross-lingual & mult. CoVe + auto & 9.28 &	 59.25 & {53.89} &		19.25		 \\
                  Cross-lingual & ELMo &  0.18 &	35.98  &  35.36 & 2.53				 \\ \midrule \midrule
        \textbf{Thai}  & Embedding type &  Exact match & Domain acc. & Intent acc. & Slot F1 \\ \midrule
                 Translate train & - & \textbf{45.59} & \textbf{98.11} & \textbf{95.85} & \textbf{55.43} \\ \midrule 

         Cross-lingual & - & 0.20 & 39.36 & 39.11 &			3.44 	  \\ 
         Cross-lingual & CoVe & 	5.82 & 66.75 & 54.24 & 		8.84	  \\ 
         Cross-lingual & mult. CoVe & 15.37  &73.84 &	 66.35 & 	32.52 
		  \\ 
         Cross-lingual & mult. CoVe + auto & {20.84}  &	{81.95} & {70.70} &		{35.62} \\
    \end{tabular}
    \vspace{1em}
    \caption{Zero-shot results averaged over 5 training runs. All models were trained only on the English data. In the case of the \textit{translate train} models, the English data was automatically translated into the target language.}
    \label{tab:zero-shot-results}
\end{table*}

\paragraph{Implementation details} We implement all classification and sequence labeling models within the PyText framework \cite{Aly2018}. We train models for 20 epochs and select the model that performs best on the development set. We use the Adam optimizer \cite{Kingma2015} with a learning rate of 0.01. We use dropout of 0.3 in the BiLSTM and we set the size of the self-attention layer to 128 dimensions.

\paragraph{Evaluation}
We evaluate our models according to four metrics: Domain accuracy, which measures the accuracy of the domain classification task; intent accuracy, which measures the accuracy of identifying the correct intent; slot F1, which is the geometric mean of the slot precision and slot recall; and the exact match metric, which indicates the number of utterances for which the domain, intent, and all slots were correctly identified. Exact match is thus the strictest metric of all. We micro-average all metrics across domains.

\paragraph{Results and discussion}

Table~\ref{tbl:full-results} shows the results for all evaluated models. While we get slightly different results for the two languages, there are several consistent patterns. For Spanish, we observe that adding contextual word representations to the \textit{target only} model, consistently improves results. Not surprisingly since the ELMo embeddings were trained on a large monolingual corpus, the model that uses these embeddings outperforms all other \textit{target only} models.

If we turn to the cross-lingual models for Spanish, the results indicate that the translation method works well for domain and intent classification but less so for slot detection, presumably due to noisy projection of the slot annotations. For slot detection, we get the best results using the ELMo embeddings which outperform the XLU embeddings as well as the bidirectional MT encoder in terms of exact match and slot F$_1$. Similarly as in the \textit{target only} setting, the model with multilingual CoVe embeddings combined with the autoencoder performs worse than the other CoVe encoders. Overall, however, the choice of embeddings seems to have only a relatively small impact on the performance of the cross-lingual models. Importantly, however, we see improvements across all metrics as compared to training only on the target language data.

We observe similar results for Thai. The translation approach again yields the worst results for slot detection and we again see a consistent improvement from cross-lingual training as compared to training only on Thai data. And we again only observe small differences depending on the type of embeddings in the cross-lingual training scenario, but in this case, the models with the multilingual CoVe encoders outperform the model with the monolingual encoder.

Table~\ref{tbl:full-results} also shows the results for English. Not surprisingly, since we have an order of magnitude more data for English, the model trained and tested on English data still performs better than any of the evaluated methods for the other two languages. However, the gap between the numbers for English and the numbers for the other two languages does get considerably smaller for the cross-lingual models. Prima facie, the results also indicate that the models perform better for Thai than for Spanish. However, this is potentially an artifact of the data. As we mentioned above, we had to discard some of the Thai utterances for which the annotations disagreed with the annotations of their English translations and it is possible that we discarded a disproportionate number of more complex utterances which in return made parsing the Thai utterances easier. 

In summary, the results from both languages suggest that pre-trained word representations as well as cross-lingual training improve results over training only on target language data without any pre-trained embeddings. The choice of embeddings, however, seems to matter less, and the overall performance seems to depend only very little on whether we use dynamic or static word representations or whether we use monolingual or bilingual word representations. 

However, interestingly, for Spanish for which we compared more types of word representations than for Thai, the cross-lingual model with monolingual ELMo embeddings provided the best results. This potentially indicates that the benefit of cross-lingual training comes from sharing the biLSTM layer or the CRF layer and that embedding the tokens of the high-resource and the low-resource language in a similar space is not as important. At the same time, considering that we are getting relatively good results for both languages if we only train on the target language data, the potential of cross-lingual training might be limited in this case. To investigate whether the embedding type matters in more extreme low-resource scenarios, we also performed a series of zero-shot and low-resource experiments, which we describe in the next section.

\begin{figure*}
    \centering
    \includegraphics[width=.7\textwidth]{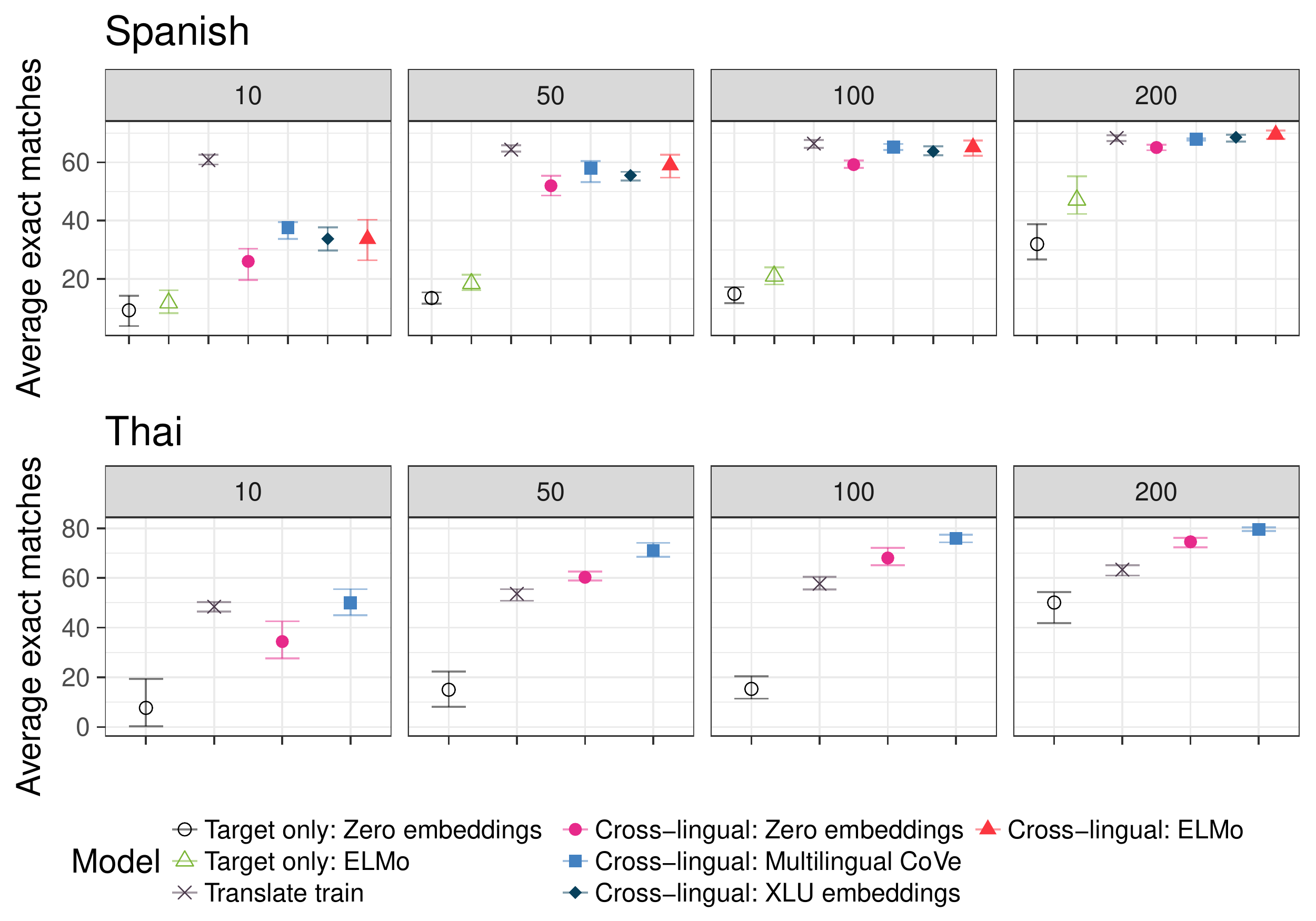} \\
    \caption{Results for different training set sizes.  The top and the bottom of the error bars correspond to the highest and lowest value of the exact match metric among the 10 runs.}
    \label{fig:learning-curve}
\end{figure*}

\section{Zero-shot learning and learning curves}

As mentioned in the previous section, from the results on the full data, it is not entirely clear whether there is an advantage of using cross-lingual embeddings. We therefore conducted additional experiments with even smaller training sets in the target language: the case where we no annotated data in the target language exists (zero-shot) and the case where a very limited amount of training data in the target language exists. If there is no advantage of using cross-lingual embeddings over using monolingual embeddings, we expect to see similar results for all models independent of the training data size. On the other hand, if the multilingual CoVe encoder actually embeds phrases with similar meanings in the two languages in a similar vector space, we would expect the model with multilingual CoVe embeddings to perform much better in the zero-shot and very low-resource scenarios than any of the models with monolingual embeddings. Further, we can also investigate whether there is an advantage of translating the training data over other methods in extremely low-resource scenarios.

\paragraph{Experiments}

We used the same models with the same parameters as in the previous section. In the zero-shot case, we only use English data for training and model selection. For the learning curve experiments, we sample 10, 50, 100, or 200 utterances from each domain for the target language for training and model selection and upsample the target language data so that it roughly matches the size of the English data. For the zero-shot results, we present the average numbers across 5 runs. For the learning curve experiments, since we introduced another random factor by randomly sampling the training and model selection data, we repeat this process 10 times and report the average as well as the minimum and maximum values of the exact match metric for these experiments.

\paragraph{Results and discussion}

Table~\ref{tab:zero-shot-results} shows the zero-shot results. These results indicate that using the multilingual CoVe embeddings works better than not using any encoder embeddings or using monolingual CoVe embeddings. This is true for the sentence-level domain and intent classification tasks as well as for slot detection. The Spanish results also suggest that in the zero-shot case, the multilingual encoder embeddings lead to better results than the XLU embeddings. However, also the models that use cross-lingual embeddings perform very poorly and contrary to the case where we have some training data in the target language, in the zero-shot scenario, the translation method works considerably better than any other of the transfer methods.

The results for different training set sizes are shown in Figure~\ref{fig:learning-curve}. These results generally confirm the patterns that we observed in the experiments with all available training data: cross-lingual training improves the results over training only on the target language (to a much bigger extent when there is much less target language training data available) and using pre-trained word representations leads to further improvements. We further observe that cross-lingual learning seems to lead to much more stable training which can be seen in the much smaller ranges of results as compared to the  models trained only on the target language. Also in the extremely low-resource scenarios, the choice of embeddings seems to have very little effect. Lastly, as these plots show, the translation approach works better when there is very little training data available but the performance quickly plateaus and once there are are several hundred target language training examples available, joint training on multiple languages leads to better results.

Considering all results together, we find a consistent advantage of using cross-lingual training across all languages, training set sizes and embedding types. We further observe that the choice of embedding type has little effect as long as some form of pre-trained embeddings is being used. These facts together suggest that the main advantage of cross-lingual training comes from sharing the biLSTM and CRF layer across languages. This is in line with the results by \citet{DeLhoneux2018} who found for cross-lingual training of dependency parsers, sharing the MLP layer for parser decisions improved results for all language pairs that they considered, whereas sharing of lower-level parameters only led to improvements for a limited set of language pairs. 

\section{Related work}

\paragraph{Cross-lingual sequence labeling}
The task of cross-lingual and multilingual sequence labeling has gained a lot of attention recently. \citet{Yang2017} used shared character embeddings for cross-lingual transfer, and \citet{Lin2018} used shared character and sentence embeddings that were trained in a multitask setting for part-of-speech tagging and named entity recognition.
\citet{Upadhyay2018} used cross-lingual embeddings for training multilingual slot-filling systems. 
\citet{Xie2018} used a similar model for NER but they first ``translated'' the high-resource training data by replacing each token with the token in the target language that was closest in vector space, and they further used character embeddings and a self-attention mechanism. \citet{YuX2018} investigated using character-based language 
models for NER in several languages but did not do any cross-lingual learning. \citet{Plank2018} used cross-lingual embeddings, projected annotations, and dictionaries for zero-shot cross-lingual part-of-speech tagging. 

\paragraph{Cross-lingual sentence representations}
Recently, there was also a lot of work of using cross-lingual sequence encoders for sentence classifications using either multilingual MT encoders similar to ours (e.g., \citealp{Eriguchi2018,YuK2018,Singla2018}) or training encoders and then aligning their vector spaces after pre-training \citep{Conneau2018, Schuster2019}. Even more recently, \citet{Conneau2019} and \citet{Mulcaire2019} showed that it is also possible to directly train contextual word representations jointly on multiple languages. 

\paragraph{Cross-lingual transfer for other tasks}
Apart from tasks such as slot filling and NER, cross-lingual transfer learning has also been investigated a lot for syntactic tasks, and in particular for part-of-speech tagging and dependency parsing. Early work trained part-of-speech taggers for individual languages and then trained delexicalized dependency parsers (e.g., \citealp{Zeman2008,McDonald2011}). Further, a lot of syntactic and semantic parsing models recently successfully incorporated parameter sharing for training parsers in closely related languges \citep{Duong2015, Ammar2016, Susanto2017, Smith2018, DeLhoneux2018}. In the domain of dialog managers,  \citet{Mrksic2017}  and \citet{Chen2018} presented methods for cross-lingual transfer for dialog state  tracking.

\section{Conclusion and future work}

In this paper, we presented a new multilingual intent and slot filling data set for task oriented dialog of around 57,000 utterances and evaluated the performance of different methods for cross-lingual transfer learning, including a novel method using cross-lingual contextual word representations. For both investigated languages, we consistently found that cross-lingual learning improves results as compared to only training on limited amounts of target language data, and our results suggest that 
the choice of multilingual or monolingual embeddings has only a small impact on the overall performance.

Despite the range of models that we considered in this paper, we only scratched at the surface of possible cross-lingual (embedding) models, and hence there are many future directions of this work. First, except for the Spanish ELMo embeddings,  we did not use any character embeddings in any of our experiments or models. This presumably makes sense for the English-Thai transfer learning case since these two languages use different writing systems but given the results by \citet{Lin2018} and \citet{Yang2017}, we would expect additional improvements by sharing character embeddings for languages with similar writing systems.

Second, one could try to include a specific learning objective to embed translations into a similar vector space as used by \citet{YuK2018} and \citet{Conneau2018} for multilingual sentence representations.

As yet another extension, one could combine the approaches of multilingual CoVe embeddings and monolingual ELMo (or BERT, \citealp{Devlin2018}) embeddings and jointly train an encoder with a language model and an MT objective, which would potentially combine the benefit of training a model on large monolingual corpora while at the same time aligning the vector spaces of the two languages. A similar approach worked well for cross-lingual NLI inference on the XNLI data set \citep{Conneau2018} as well as for unsupervised machine translation \citep{Conneau2019}.

We hope that our data set will facilitate research in these directions and ultimately lead to improved natural language understanding models for low-resource languages.

\section*{Acknowledgements}

We thank the three anonymous reviewers for their thoughtful comments.
We also thank Luke Zettlemoyer and Christopher Manning for valuable
feedback throughout this project, and Maria Sumner and Neghesti Tinsew for 
their help with collecting the annotations and improving their consistency.
\bibliography{references}

\begin{thebibliography}{45}
\expandafter\ifx\csname natexlab\endcsname\relax\def\natexlab#1{#1}\fi

\bibitem[{Aly et~al.(2018)Aly, Lakhotia, Zhao, Mohit, Oguz, Arora, Gupta,
  Dewan, Nelson-Lindall, and Shah}]{Aly2018}
Ahmed Aly, Kushal Lakhotia, Shicong Zhao, Mrinal Mohit, Barlas Oguz, Abhinav
  Arora, Sonal Gupta, Christopher Dewan, Stef Nelson-Lindall, and Rushin Shah.
  2018.
\newblock \href {https://arxiv.org/pdf/1812.08729.pdf} {{PyText: A} seamless
  path from {NLP} research to production}.
\newblock ArXiv preprint.

\bibitem[{Ammar et~al.(2016)Ammar, Mulcaire, Ballesteros, Dyer, and
  Smith}]{Ammar2016}
Waleed Ammar, George Mulcaire, Miguel Ballesteros, Chris Dyer, and Noah~A.
  Smith. 2016.
\newblock \href {https://www.transacl.org/ojs/index.php/tacl/article/view/892}
  {Many languages, one parser}.
\newblock \emph{Transactions of the Association for Computational Linguistics},
  4:431--444.

\bibitem[{Cettolo et~al.(2012)Cettolo, Girardi, and Federico}]{Cettolo2012}
Mauro Cettolo, Christian Girardi, and Marcello Federico. 2012.
\newblock \href {http://www.mt-archive.info/EAMT-2012-Cettolo.pdf} {Wit$^3$:
  {W}eb inventory of transcribed and translated talks}.
\newblock In \emph{Proceedings of the 16$^{th}$ Conference of the European
  Association for Machine Translation (EAMT)}, Trento, Italy.

\bibitem[{Che et~al.(2018)Che, Liu, Wang, Zheng, and Liu}]{Che2018}
Wanxiang Che, Yijia Liu, Yuxuan Wang, Bo~Zheng, and Ting Liu. 2018.
\newblock \href {http://www.aclweb.org/anthology/K18-2005} {Towards better {UD}
  parsing: Deep contextualized word embeddings, ensemble, and treebank
  concatenation}.
\newblock In \emph{Proceedings of the {CoNLL} 2018 Shared Task: {M}ultilingual
  Parsing from Raw Text to Universal Dependencies}.

\bibitem[{Chen et~al.(2018)Chen, Chen, Su, Wang, Yu, Yan, and Wang}]{Chen2018}
Wenhu Chen, Jianshu Chen, Yu~Su, Xin Wang, Dong Yu, Xifeng Yan, and
  William~Yang Wang. 2018.
\newblock \href {http://www.aclweb.org/anthology/D18-1038} {{XL-NBT}: {A}
  cross-lingual neural belief tracking framework}.
\newblock In \emph{Proceedings of the 2018 Conference on Empirical Methods in
  Natural Language Processing (EMNLP 2018)}, pages 414--424.

\bibitem[{Conneau et~al.(2017)Conneau, Lample, Ranzato, Denoyer, and
  J{\'{e}}gou}]{Conneau2017}
Alexis Conneau, Guillaume Lample, Marc'Aurelio Ranzato, Ludovic Denoyer, and
  Herv{\'{e}} J{\'{e}}gou. 2017.
\newblock \href {https://arxiv.org/abs/1710.04087} {Word translation without
  parallel data}.
\newblock In \emph{Proceedings of ICLR 2018}.

\bibitem[{Conneau et~al.(2018)Conneau, Lample, Rinott, Schwenk, Stoyanov,
  Williams, and Bowman}]{Conneau2018}
Alexis Conneau, Guillaume Lample, Ruty Rinott, Holger Schwenk, Ves Stoyanov,
  Adina Williams, and Samuel~R. Bowman. 2018.
\newblock \href {http://www.aclweb.org/anthology/D18-1269} {{XNLI}:
  {E}valuating cross-lingual sentence representations}.
\newblock In \emph{Proceedings of the 2018 Conference on Empirical Methods in
  Natural Language Processing (EMNLP 2018)}, pages 2475--2485.

\bibitem[{Devlin et~al.(2018)Devlin, Chang, Lee, and Toutanova}]{Devlin2018}
Jacob Devlin, Ming-Wei Chang, Kenton Lee, and Kristina Toutanova. 2018.
\newblock \href {http://arxiv.org/abs/1810.04805} {{BERT}: {P}re-training of
  deep bidirectional transformers for language understanding}.
\newblock ArXiv preprint.

\bibitem[{Dozat et~al.(2017)Dozat, Qi, and Manning}]{Dozat2017}
Timothy Dozat, Peng Qi, and Christopher~D. Manning. 2017.
\newblock \href {http://aclweb.org/anthology/K17-3002} {{S}tanford's
  graph-based neural dependency parser at the {CoNLL} 2017 shared task}.
\newblock In \emph{Proceedings of the CoNLL 2017 Shared Task: Multilingual
  Parsing from Raw Text to Universal Dependencies}.

\bibitem[{Duong et~al.(2015)Duong, Cohn, Bird, and Cook}]{Duong2015}
Long Duong, Trevor Cohn, Steven Bird, and Paul Cook. 2015.
\newblock \href {http://www.aclweb.org/anthology/P15-2139} {Low resource
  dependency parsing: {C}ross-lingual parameter sharing in a neural network
  parser}.
\newblock In \emph{Proceedings of the 53rd Annual Meeting of the Association
  for Computational Linguistics and the 7th International Joint Conference on
  Natural Language Processing (ACL 2015)}, pages 845--850.

\bibitem[{Eriguchi et~al.(2018)Eriguchi, Johnson, Firat, Kazawa, and
  Macherey}]{Eriguchi2018}
Akiko Eriguchi, Melvin Johnson, Orhan Firat, Hideto Kazawa, and Wolfgang
  Macherey. 2018.
\newblock \href {https://arxiv.org/abs/1809.04686} {Zero-shot cross-lingual
  classification using multilingual neural machine translation}.
\newblock ArXiv preprint.

\bibitem[{Fares et~al.(2017)Fares, Kutuzov, Oepen, and Velldal}]{Fares2017}
Murhaf Fares, Andrey Kutuzov, Stephan Oepen, and Erik Velldal. 2017.
\newblock \href {http://aclweb.org/anthology/W17-0237} {Word vectors, reuse,
  and replicability: {T}owards a community repository of large-text resources}.
\newblock In \emph{Proceedings of the 21st Nordic Conference on Computational
  Linguistics}, pages 271--276.

\bibitem[{Gehring et~al.(2016)Gehring, Auli, Grangier, and
  Dauphin}]{Gehring2016}
Jonas Gehring, Michael Auli, David Grangier, and Yann~N Dauphin. 2016.
\newblock \href {https://arxiv.org/abs/1611.02344} {{A Convolutional Encoder
  Model for Neural Machine Translation}}.
\newblock ArXiv preprint.

\bibitem[{Gehring et~al.(2017)Gehring, Auli, Grangier, Yarats, and
  Dauphin}]{Gehring2017}
Jonas Gehring, Michael Auli, David Grangier, Denis Yarats, and Yann~N Dauphin.
  2017.
\newblock \href {https://arxiv.org/abs/1705.03122} {{Convolutional Sequence to
  Sequence Learning}}.
\newblock ArXiv preprint.

\bibitem[{Kingma and Ba(2015)}]{Kingma2015}
Diederik~P. Kingma and Jimmy~Lei Ba. 2015.
\newblock \href {http://arxiv.org/abs/1412.6980} {Adam: {A} method for
  stochastic optimization}.
\newblock In \emph{Proceedings of ICLR 2015}.

\bibitem[{Koehn(2005)}]{Koehn2005}
Philipp Koehn. 2005.
\newblock \href
  {http://homepages.inf.ed.ac.uk/pkoehn/publications/europarl-mtsummit05.pdf}
  {Europarl: {A} parallel corpus for statistical machine translation}.
\newblock In \emph{MT summit}.

\bibitem[{Lample et~al.(2016)Lample, Ballesteros, Subramanian, Kawakami, and
  Dyer}]{lamplener}
Guillaume Lample, Miguel Ballesteros, Sandeep Subramanian, Kazuya Kawakami, and
  Chris Dyer. 2016.
\newblock \href {http://www.aclweb.org/anthology/N16-1030} {Neural
  architectures for named entity recognition}.
\newblock In \emph{Proceedings of the 2016 Conference of the North American
  Chapter of the Association for Computational Linguistics: Human Language
  Technologies (NAACL 2016)}, pages 260--270.

\bibitem[{Lample and Conneau(2019)}]{Conneau2019}
Guillaume Lample and Alexis Conneau. 2019.
\newblock \href {https://arxiv.org/pdf/1901.07291.pdf} {Cross-lingual language
  model pretraining}.
\newblock ArXiv preprint.

\bibitem[{de~Lhoneux et~al.(2018)de~Lhoneux, Bjerva, Augenstein, and
  S{\o}gaard}]{DeLhoneux2018}
Miryam de~Lhoneux, Johannes Bjerva, Isabelle Augenstein, and Anders S{\o}gaard.
  2018.
\newblock \href {http://www.aclweb.org/anthology/D18-1543} {Parameter sharing
  between dependency parsers for related languages}.
\newblock In \emph{Proceedings of the 2018 Conference on Empirical Methods in
  Natural Language Processing (EMNLP 2018)}, pages 4992--4997.

\bibitem[{L'Hostis et~al.(2016)L'Hostis, Grangier, and Auli}]{Lhostis2016}
Gurvan L'Hostis, David Grangier, and Michael Auli. 2016.
\newblock \href {https://arxiv.org/abs/1610.00072} {Vocabulary selection
  strategies for neural machine translation}.
\newblock ArXiv preprint.

\bibitem[{Lin et~al.(2018)Lin, Yang, Stoyanov, and Ji}]{Lin2018}
Ying Lin, Shengqi Yang, Veselin Stoyanov, and Heng Ji. 2018.
\newblock \href {http://www.aclweb.org/anthology/P18-1074} {A multi-lingual
  multi-task architecture for low-resource sequence labeling}.
\newblock In \emph{Proceedings of the 56th Annual Meeting of the Association
  for Computational Linguistics (ACL 2018)}, pages 799--809.

\bibitem[{Lin et~al.(2017)Lin, Feng, dos Santos, Yu, Xiang, Zhou, and
  Bengio}]{Lin2017}
Zhouhan Lin, Minwei Feng, Cicero~Nogueira dos Santos, Mo~Yu, Bing Xiang, Bowen
  Zhou, and Yoshua Bengio. 2017.
\newblock \href {https://doi.org/10.1109/CVPR.2016.90} {A structured
  self-attentive sentence embedding}.
\newblock In \emph{Proceedings of ICLR 2017}.

\bibitem[{Lison et~al.(2018)Lison, Tiedemann, and Kouylekov}]{Lison2018}
Pierre Lison, J{\"o}rg Tiedemann, and Milen Kouylekov. 2018.
\newblock \href {http://www.lrec-conf.org/proceedings/lrec2018/pdf/294.pdf}
  {Open{S}ubtitles2018: {S}tatistical rescoring of sentence alignments in
  large, noisy parallel corpora}.
\newblock In \emph{Proceedings of the 11th edition of the Language Resources
  and Evaluation Conference (LREC 2018)}, pages 1364--1369.

\bibitem[{Liu and Lane(2016)}]{Liu2016AttentionBasedRN}
Bing Liu and Ian Lane. 2016.
\newblock \href {https://arxiv.org/abs/1609.01454} {Attention-based recurrent
  neural network models for joint intent detection and slot filling}.
\newblock In \emph{Proceedings of Interspeech 2016}.

\bibitem[{Luong et~al.(2015)Luong, Pham, and Manning}]{Luong2015}
Minh-Thang Luong, Hieu Pham, and Christopher~D. Manning. 2015.
\newblock \href {http://aclweb.org/anthology/D15-1166} {Effective approaches to
  attention-based neural machine translation}.
\newblock In \emph{Proceedings of the 2015 Conference on Empirical Methods in
  Natural Language Processing (EMNLP 2015)}, pages 1412--1421.

\bibitem[{{McCann} et~al.(2017){McCann}, Bradbury, and Socher}]{McCann2017}
Bryan {McCann}, James Bradbury, and Richard Socher. 2017.
\newblock \href {https://arxiv.org/abs/1708.00107} {Learned in translation:
  {C}ontextualized word vectors}.
\newblock In \emph{Proceedings of the 31st Conference on Neural Information
  Processing Systems (NIPS 2017)}.

\bibitem[{McDonald et~al.(2011)McDonald, Petrov, and Hall}]{McDonald2011}
Ryan McDonald, Slav Petrov, and Keith Hall. 2011.
\newblock \href {http://www.aclweb.org/anthology/D11-1006} {Multi-source
  transfer of delexicalized dependency parsers}.
\newblock In \emph{Proceedings of the 2011 Conference on Empirical Methods in
  Natural Language Processing (EMNLP 2011)}, pages 62--72.

\bibitem[{Mesnil et~al.(2013)Mesnil, He, Deng, and Bengio}]{mesnil2013}
Gr\'egoire Mesnil, Xiaodong He, Li~Deng, and Yoshua Bengio. 2013.
\newblock \href
  {https://www.iro.umontreal.ca/~lisa/pointeurs/RNNSpokenLanguage2013.pdf}
  {Investigation of recurrent-neural-network architectures and learning methods
  for spoken language understanding}.
\newblock In \emph{Proceedings of Interspeech 2013}.

\bibitem[{Mrk\v{s}i\'{c} et~al.(2017)Mrk\v{s}i\'{c}, Vuli\'{c}, S\'{e}aghdha,
  Leviant, Reichart, Ga\v{s}i\'{c}, Korhonen, and Young}]{Mrksic2017}
Nikola Mrk\v{s}i\'{c}, Ivan Vuli\'{c}, Diarmuid~\'{O} S\'{e}aghdha, Ira
  Leviant, Roi Reichart, Milica Ga\v{s}i\'{c}, Anna Korhonen, and Steve Young.
  2017.
\newblock \href {https://doi.org/10.1162/tacl\_a\_00063} {Semantic
  specialization of distributional word vector spaces using monolingual and
  cross-lingual constraints}.
\newblock \emph{Transactions of the Association for Computational Linguistics},
  5:309--324.

\bibitem[{Mulcaire et~al.(2019)Mulcaire, Kasai, and Smith}]{Mulcaire2019}
Phoebe Mulcaire, Jungo Kasai, and Noah~A. Smith. 2019.
\newblock Polyglot contextual representations improve crosslingual transfer.
\newblock In \emph{Proceedings of the 2019 Conference of the North American
  Chapter of the Association for Computational Linguistics: Human Language
  Technologies (NAACL 2019)}.

\bibitem[{Peters et~al.(2018)Peters, Neumann, Iyyer, Gardner, Clark, Lee, and
  Zettlemoyer}]{Peters2018}
Matthew Peters, Mark Neumann, Mohit Iyyer, Matt Gardner, Christopher Clark,
  Kenton Lee, and Luke Zettlemoyer. 2018.
\newblock \href {https://doi.org/10.18653/v1/N18-1202} {Deep contextualized
  word representations}.
\newblock In \emph{Proceedings of the 2018 Conference of the North American
  Chapter of the Association for Computational Linguistics: Human Language
  Technologies (NAACL 2018)}.

\bibitem[{Plank and Agi{\'c}(2018)}]{Plank2018}
Barbara Plank and {\v{Z}}eljko Agi{\'c}. 2018.
\newblock \href {https://www.aclweb.org/anthology/D18-1061} {Distant
  supervision from disparate sources for low-resource part-of-speech tagging}.
\newblock In \emph{Proceedings of the 2018 Conference on Empirical Methods in
  Natural Language Processing}, pages 614--620.

\bibitem[{Ruder et~al.(2017)Ruder, Vuli{\'{c}}, and S{\o}gaard}]{Ruder2017}
Sebastian Ruder, Ivan Vuli{\'{c}}, and Anders S{\o}gaard. 2017.
\newblock \href {http://arxiv.org/abs/1706.04902} {A survey of cross-lingual
  word embedding models}.
\newblock ArXiv preprint.

\bibitem[{Schuster et~al.(2019)Schuster, Ram, Barzilay, and
  Globerson}]{Schuster2019}
Tal Schuster, Ori Ram, Regina Barzilay, and Amir Globerson. 2019.
\newblock Cross-lingual alignment of contextual word embeddings, with
  applications to zero-shot dependency parsing.
\newblock In \emph{Proceedings of the 2019 Conference of the North American
  Chapter of the Association for Computational Linguistics: Human Language
  Technologies (NAACL 2019)}.

\bibitem[{Singla et~al.(2018)Singla, Can, and Narayanan}]{Singla2018}
Karan Singla, Dogan Can, and Shrikanth Narayanan. 2018.
\newblock \href {http://www.aclweb.org/anthology/P18-2035} {A multi-task
  approach to learning multilingual representations}.
\newblock In \emph{Proceedings of the 56th Annual Meeting of the Association
  for Computational Linguistics (ACL 2018)}, pages 214--220.

\bibitem[{Smith et~al.(2018)Smith, Bohnet, and {de Lhoneux}}]{Smith2018}
Aaron Smith, Bernd Bohnet, and Miryam {de Lhoneux}. 2018.
\newblock \href {https://doi.org/10.18653/v1/K18-2011} {82 treebanks, 34
  models: {U}niversal {D}ependency parsing with multi-treebank models}.
\newblock In \emph{Proceedings of the CoNLL 2018 Shared Task: Multilingual
  Parsing from Raw Text to Universal Dependencies}.

\bibitem[{Susanto and Lu(2017)}]{Susanto2017}
Raymond~Hendy Susanto and Wei Lu. 2017.
\newblock \href {http://aclweb.org/anthology/P17-2007} {Neural architectures
  for multilingual semantic parsing}.
\newblock In \emph{Proceedings of the 55th Annual Meeting ofthe Association for
  Computational Linguistics (ACL 2017)}, pages 38--44.

\bibitem[{Upadhyay et~al.(2018)Upadhyay, Faruqui, T{\"u}r, Hakkani-T{\"u}r, and
  Heck}]{Upadhyay2018}
Shyam Upadhyay, Manaal Faruqui, Gokhan T{\"u}r, Dilek Hakkani-T{\"u}r, and
  Larry Heck. 2018.
\newblock \href {http://shyamupa.com/papers/UFTHH18.pdf} {({A}lmost) zero-shot
  cross-lingual spoken language understanding}.
\newblock In \emph{Proceedings of the IEEE ICASSP 2018}.

\bibitem[{Vu(2016)}]{Vu2016}
Ngoc~Thang Vu. 2016.
\newblock \href {https://arxiv.org/abs/1606.07783} {Sequential convolutional
  neural networks for slot filling in spoken language understanding}.
\newblock In \emph{Proceedings of Interspeech 2016}, pages 3250--3254.

\bibitem[{Xie et~al.(2018)Xie, Yang, Neubig, Smith, and Carbonell}]{Xie2018}
Jiateng Xie, Zhilin Yang, Graham Neubig, Noah~A. Smith, and Jaime Carbonell.
  2018.
\newblock \href {http://www.aclweb.org/anthology/D18-1034} {Neural
  cross-lingual named entity recognition with minimal resources}.
\newblock In \emph{Proceedings of the 2018 Conference on Empirical Methods in
  Natural Language Processing (EMNLP 2018)}, pages 369--379.

\bibitem[{Yang et~al.(2017)Yang, Salakhutdinov, and Cohen}]{Yang2017}
Zhilin Yang, Ruslan Salakhutdinov, and William~W. Cohen. 2017.
\newblock \href {https://arxiv.org/abs/1703.06345} {Transfer learning for
  sequence tagging with hierarchical recurrent networks}.
\newblock In \emph{Proceedings of ICLR 2017}.

\bibitem[{Yarowsky et~al.(2001)Yarowsky, Ngai, and Wicentowski}]{Yarowsky2001}
David Yarowsky, Grace Ngai, and Richard Wicentowski. 2001.
\newblock \href {https://www.aclweb.org/anthology/H01-1035} {Inducing
  multilingual text analysis tools via robust projection across aligned
  corpora}.
\newblock In \emph{Proceedings of the First International Conference on Human
  Language Technology Research}.

\bibitem[{Yu et~al.(2018{\natexlab{a}})Yu, Li, and Oguz}]{YuK2018}
Katherin Yu, Haoran Li, and Barlas Oguz. 2018{\natexlab{a}}.
\newblock \href {http://www.aclweb.org/anthology/W18-3023} {Multilingual
  seq2seq training with similarity loss for cross-lingual document
  classification}.
\newblock In \emph{Proceedings of The Third Workshop on Representation Learning
  for NLP}, pages 175--179.

\bibitem[{Yu et~al.(2018{\natexlab{b}})Yu, Mayhew, Sammons, and Roth}]{YuX2018}
Xiaodong Yu, Stephen Mayhew, Mark Sammons, and Dan Roth. 2018{\natexlab{b}}.
\newblock \href {http://www.aclweb.org/anthology/D18-1345} {On the strength of
  character language models for multilingual named entity recognition}.
\newblock In \emph{Proceedings of the 2018 Conference on Empirical Methods in
  Natural Language Processing (EMNLP 2018)}, pages 3073--3077.

\bibitem[{Zeman and Resnik(2008)}]{Zeman2008}
Daniel Zeman and Philip Resnik. 2008.
\newblock \href {http://www.aclweb.org/anthology/I08-3008} {Cross-language
  parser adaptation between related languages}.
\newblock In \emph{Proceedings of the IJCNLP-08 Workshop on NLP for Less
  Privileged Languages}.

\end{thebibliography}
\bibliographystyle{acl_natbib}

\end{document}